\documentclass[journal,twoside,web]{IEEEtran}
\usepackage{cite}
\usepackage{amsmath,amssymb,amsfonts}
\usepackage{graphicx}
\usepackage{textcomp}
\usepackage{hyperref}
\usepackage{algpseudocode}
\usepackage{varwidth}
\usepackage{algorithm}
\usepackage{color}

\def \AsO {\mathcal{O}}

\def\BibTeX{{\rm B\kern-.05em{\sc i\kern-.025em b}\kern-.08em
    T\kern-.1667em\lower.7ex\hbox{E}\kern-.125emX}}
\begin{document}
\title{A Streaming Volumetric Image Generation Framework for Development and Evaluation of Out-of-Core Methods}
\author{Dominik Drees, and Xiaoyi Jiang, \IEEEmembership{Senior Member, IEEE}
        \thanks{D. Drees and X. Jiang are with the Faculty of Mathematics and Computer Science, University of Münster, Münster, Germany. (e-mail: xjiang@uni-muenster.de)}}

\maketitle

\begin{abstract}
Advances in 3D imaging technology in recent years have allowed for increasingly high resolution volumetric images of large specimen.
The resulting datasets of hundreds of Gigabytes in size call for new scalable and memory efficient approaches in the field of image processing, where some progress has been made already.
At the same time, quantitative evaluation of these new methods is difficult both in terms of the availability of specific data sizes and in the generation of associated ground truth data.
In this paper we present an algorithmic framework that can be used to efficiently generate test (and ground truth) volume data, optionally even in a streaming fashion.
As the proposed \textit{nested sweeps} algorithm is fast, it can be used to generate test data on demand.
We analyze the asymptotic run time of the presented algorithm and compare it experimentally to alternative approaches as well as a hypothetical best-case baseline method.
In a case study, the framework is applied to the popular VascuSynth software for vascular image generation, making it capable of efficiently producing larger-than-main memory volumes which is demonstrated by generating a trillion voxel (1TB) image.
Implementations of the presented framework are available online in the form of the modified version of Vascusynth and the code used for the experimental evaluation. In addition, the test data generation procedure has been integrated into the popular volume rendering and processing framework Voreen.
\end{abstract}


\section{Introduction}

In recent years, there have been advances in 3D imaging technologies which, for example, promise greater insights into biological processes happening at multiple spatial scales~\cite{wan2019light, zhao2020Cellular, valm2017applying}.
Extracting understanding from these datasets often requires redesign of existing methods or development of entirely new out-of-core (i.e.\ operating on larger-than-main memory data) and/or sparse (i.e.\ operating on a non-dense subset of the data) algorithms~\cite{kirst2020mapping, drees2021hierarchical, isenburg2009streaming, drees2021scalable, fang2015sparsity} due to the massive size of the resulting volumetric image datasets.
Development and especially evaluation of these new algorithms is difficult for several reasons:
(1) Creation of accurate ground truth (for example for a voxel-wise segmentation application) for existing large, high resolution real world is often infeasible.
(2) Existing real world datasets have a fixed resolution and size and the variation thereof between comparable datasets (e.g.\ from the same published database) is usually small.
As a result, experimental demonstration of run time trends for different data sizes for otherwise comparable datasets is difficult.
Additionally, practical problems of the implementation of large data methods sometimes only manifest themselves when \textit{actually} operating on large datasets:
This can mean that previous performance uncritical parts of the pipeline turn out to be the bottleneck, or that the performance of used operating system services (e.g.\ the file system) drops off due to the unusual usage pattern.
Hence, it is desirable to control the size of datasets used for development and evaluation independent of other parameters.
(3) Similarly, other domain specific parameters (e.g.\ radii of vessel structures or density of cell clusters) cannot be manipulated easily.
The cost of acquisition (4) and storing (5) large numbers of datasets to achieve statistical significance in a quantitative evaluation is high.
Problem (1) in particular also shows a bootstrapping problem:
Manual or semiautomatic creation of ground truth data for large datasets requires new, efficient labeling tools, but the development of these tools requires realistic, large data for evaluation.

Hence, in these situations it is desirable or even required to use a parametric scheme for test data generation.
The problem of high storage cost (5) can be eliminated by not storing generated volumes permanently and instead generate volumes on demand, which in turn requires the generation method to be fast itself (which is of course also desirable even if it \textit{is} feasible to store the generated data).

In the microscopic biomedical domain, there are two main groups of structures of interest: Vessels and cells.
For both, the problem of creating not necessarily large, but \textit{convincing} 3D datasets has already been studied extensively, which also highlights the need for artificial test data in general.
VascuSynth~\cite{hamarneh2010vascusynth} is a popular tool (see \cite{mou2021cs2,zeng2018automatic,titinunt2019vesselnet,wang2020tensorcut,koonjoo2021boosting}) for synthesis of arterial vessel trees, corresponding noisy volume data and segmentation based on oxygen demand maps.
The software implementation \cite{jassi2011vascusynth}, however, does not support out-of-core volume generation since it operates completely in memory.
VascuSynth was also used as the basis for the generation of pulmonary CT phantoms~\cite{jimenez2016automatic}.
Other works targeting volumetric vessel image generation model the tree shape based on tissue metabolism~\cite{schneider2012tissue} or experimental data~\cite{merrem2017computational}.
Another line of work targeting the generation of realistic cell models exists with focus on cell images with realistic shape and texture~\cite{svoboda2009generation, svoboda2017mitogen}, data based modeling~\cite{murphy2016building}, modeling of sub-cellular structures~\cite{murphy2012cellorganizer} and filopodia~\cite{sorokin2018filogen}, as well as cell population simulation~\cite{rajaram2012simucell, svoboda2017mitogen}.
There are also specialized works for the simulation of cells in colon tissue~\cite{svoboda2011colon} and of tubular network formation \cite{svoboda2018tubular}.
Modeling and generation of neurons~\cite{cuntz2010neuronal,torben2014context,palombo2019brain} is of special interest due to the branched nature of dendrites and axons.
Of note are also recent generative adversarial network based approaches for generation of cell shapes~\cite{wiesner2019gan} and vessel geometry~\cite{wolterink2018gan}.

While these works have proven popular and useful for algorithm evaluation and validation (e.g.\ \cite{zhao2018automatic, isensee2021nnu, castilla2019filopodia, gilad2019unsupervised}) they are not specifically designed for large test data generation. As an example, Stegmaier et al.~\cite{stegmaier2014fast} used the work of Svoboda et al.~\cite{svoboda2009generation} for quantitative evaluation, but only generated and used small datasets (of sizes in the lower Megabyte range).
Their evaluation including larger datasets (more than a Gigabyte in size) was restricted to subjective quality measure without ground truth evaluation.

Voxelization techniques in the field of volumetric constructive solid geometry~\cite{pasko1995frep} (where the focus of method development is usually interactivity rather than generation of trillion voxel images) in principle reduce the memory requirements by generating an output volume slice-by-slice on GPU~\cite{liao2002csggpu} or FPGA~\cite{koc2004fpga}, but do not solve scalability problems introduced by a large number of primitives.

\begin{figure}[t]
  \includegraphics[width=\columnwidth]{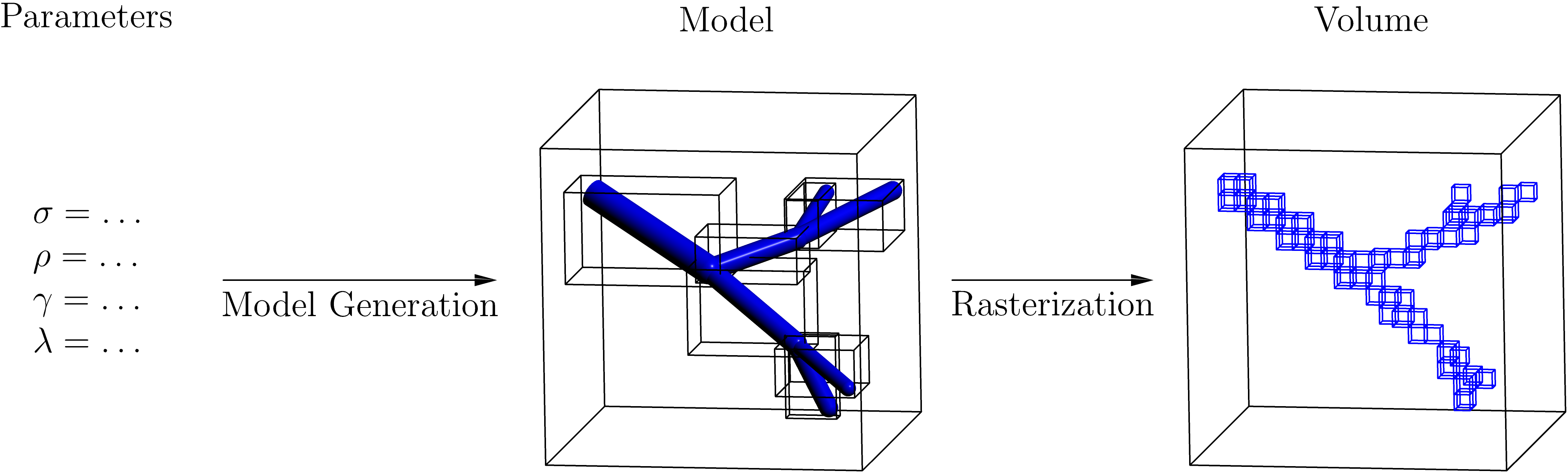}
  \caption{
  The abstract volumetric image generation pipeline used by popular data synthesis approaches (e.g.\ \cite{hamarneh2010vascusynth,svoboda2017mitogen}) and assumed in this paper:
  First, a problem specific algorithm is used to generate an intermediate model of the data based on a set of input parameters, which is then rasterized (including, for example, texture and noise generation) to obtain a volumetric image.
  Although in some sense also image specific, the intermediate model can often be expressed as a set of discrete components with implicitly defined axis-aligned bounding boxes which is exploited in this paper to present a highly efficient and out-of-core data compatible drop-in replacement for the rasterization stage of these algorithms.
}
\label{fig:motivation}
\end{figure}

In this paper, we tackle the problems (1)-(5) described above and present an algorithmic framework for efficient generation of large test data.
The framework is flexible and can be applied to freely implement domain-specific data generation schemes based on, for example, rasterization of geometric primitives and voxel-value modification.
This fits with the popular pipeline design consisting of a model generation and a subsequent rasterization stage used by many approaches~\cite{hamarneh2010vascusynth,svoboda2017mitogen} (see \autoref{fig:motivation}).
Ground truth data for segmentation algorithms can be generated in the same way in conjunction with the input data.

Given a set of primitives with associated bounding boxes (for details see \autoref{sec:problem_definition}), the run time of the \textit{nested sweeps} algorithm, the core of the proposed framework, is linear in the output volume size, total bounding box volume and log-linear in the number of primitives (under the assumption of a reasonable real world scenario).
This is proven by analysis of the algorithm and also shown experimentally, where the exact characteristics of the algorithm is compared to other methods on a hard disk drive (HDD) and a solid state drive (SSD).
The proposed framework is employed in the widely used volume rendering and processing framework Voreen~\cite{meyer2009voreen} to generate artificial vessel and cell datasets and as such was already used for evaluation of the hierarchical framework to apply the random walker method to out-of-core volumes~\cite{drees2021hierarchical}.
Similarly, as shown in a case study in this paper, it is possible to extend the popular Vascusynth method \cite{hamarneh2010vascusynth} for synthesis of volumetric images using our framework and thus enable it to efficiently generate larger-than-main memory datasets.
We are confident that the same is true for other vessel~\cite{merrem2017computational,wolterink2018gan} or cell-based~\cite{svoboda2017mitogen,sorokin2018filogen,palombo2019brain} methods.
Other than test data generation, the presented nested sweeps algorithm can also be used in the pipeline of other methods.
As an example, it can replace the nested interval trees in our previous work on vessel topology and geometry extraction~\cite{drees2021scalable} for a run time performance boost.

Our contributions can be summarized as follows:
\begin{itemize}
  \item A general framework for large volumetric test data generation is introduced.
  \item The novel and highly efficient nested sweeps algorithm for streaming of parametric data models rasterization is presented.
  \item The popular data generation method VascuSynth is augmented with support for out-of-core data using this framework.
\end{itemize}

The remainder of this paper is structured as follows.
In \autoref{sec:problem_definition}, we describe the framework both formally and in practical aspects before, in \autoref{sec:method}, introducing and analyzing two simple and a more advanced improved rasterization method that is the core of the framework.
In \autoref{sec:experiments} a number of experiments to determine the run time behavior under different real world conditions are presented and discussed.
We present the results of augmenting the VascuSynth software~\cite{jassi2011vascusynth} with support for generating larger-than-main memory volumes in \autoref{sec:case_study}. Finally, some discussion concludes the paper.

\section{Framework}
\label{sec:problem_definition}

This section presents the proposed framework in two parts:
First, it is discussed along with its interplay with concrete data generation methods (e.g.\ \cite{hamarneh2010vascusynth, svoboda2017mitogen}).
Then the framework is described formally, including the resulting rasterization problem that is solved by the methods presented in \autoref{sec:method}.

\subsection{Framework Description and Practical Considerations}
As described in the introduction, this paper presents a framework for development and adaption of volumetric test data generation methods which allows for efficient generation of extremely large and in particular larger-than-main memory datasets.
At the coarsest level, the framework dictates a two-stage pipeline structure described in \autoref{fig:motivation}.
Concretely, test data generation methods must produce an intermediate (for example parametric) model (\textit{Model Generation} stage) which can then be rasterized to produce a volumetric image (\textit{Rasterization} stage).

The Model Generation stage is of course highly domain specific and in this framework only loosely constrained by its output:
For efficient processing, the intermediate model should be split up into a (potentially large) number of smaller components.
For vascular tree generation this is the case, as the vascular tree is for example described as a set of cylinders~\cite{hamarneh2010vascusynth} or spheres~\cite{wolterink2018gan}.
In cell images individual cells are usually small and can thus form individual components.
Implicitly, each of these components defines a minimal axis-aligned bounding box, which is simply the smallest axis-aligned box that encompasses the component.
Then, each component is modeled as a function defined on its bounding box which can be sampled to generate a rasterized version of the primitive.
For vascular tree generation this function could simply output 1 if a point is within the cylinder and 0 otherwise, or only paint a bright vessel border.
Functional representations~\cite{pasko1995frep} of shapes in general lend themselves to definition of such primitives.
Further, existing methods for procedural generation of texture and interior cell structures~\cite{svoboda2017mitogen} may be used.

With this setup, in the Rasterization stage, an efficient algorithm described in \autoref{sec:method} can be applied to generate an output volume of the desired resolution.
Optionally, the output volume can be generated in a streaming fashion, i.e.\ producing voxel values one after another (in memory, i.e.\ $z$-$y$-$x$ order) for further processing without writing the volume to disk, which may be a bottleneck (see \autoref{sec:experiments}).

\subsection{Formal Definition}

After the informal description of the framework above, this section will establish notation used in the rest of the paper and formally define the resulting algorithmic problem to be solved by rasterization algorithms (see next section).

Let $d_x$, $d_y$, $d_z$ be the desired dimension of the volume to be generated by the method.
Then, $G := \{1, \ldots, d_x\} \times \{1, \ldots, d_y\} \times \{1, \ldots, d_z\}$ is the voxel grid, on which the resulting voxel values will be defined by the function $v\colon G \rightarrow V$, where $V$ is the voxel data type.
We assume that values from $V$ can be combined using a binary operation $\otimes$ or the corresponding set operation $\bigotimes$.
In the common case $V=\mathbb{R}$, the operation $\bigotimes$ will often be a summation $\Sigma$ or a maximum operation.

Then, the whole pipeline (see Figure~\ref{fig:motivation}) can be described as the combination $r \circ m$ of two functions $m$ and $r$:

\textit{The Model Generation stage} $m\colon P \rightarrow \Omega$ maps domain specific parameters from $P$ to a set of components $O = \{O^1, \ldots, O^n\} \in \Omega$.
Each component consists of an axis aligned bounding box with coordinates ($a^i_x$, $b^i_x$), ($a^i_y$, $b^i_y$), ($a^i_z$, $b^i_z$) and function $f^i$.
The bounding box also defines the subset of the full voxel grid $\{a^i_x, \ldots, b^i_x\} \times \{a^i_y, \ldots, b^i_y\} \times \{a^i_z, \ldots, b^i_z\} = G^i \subseteq G$ on which the value function is defined: $f^i\colon G^i \rightarrow V$.
It should be noted that $f^i$ is not (necessarily) rasterized, but can be and usually is defined analytically.

\begin{figure}[t]
  \includegraphics[width=\columnwidth]{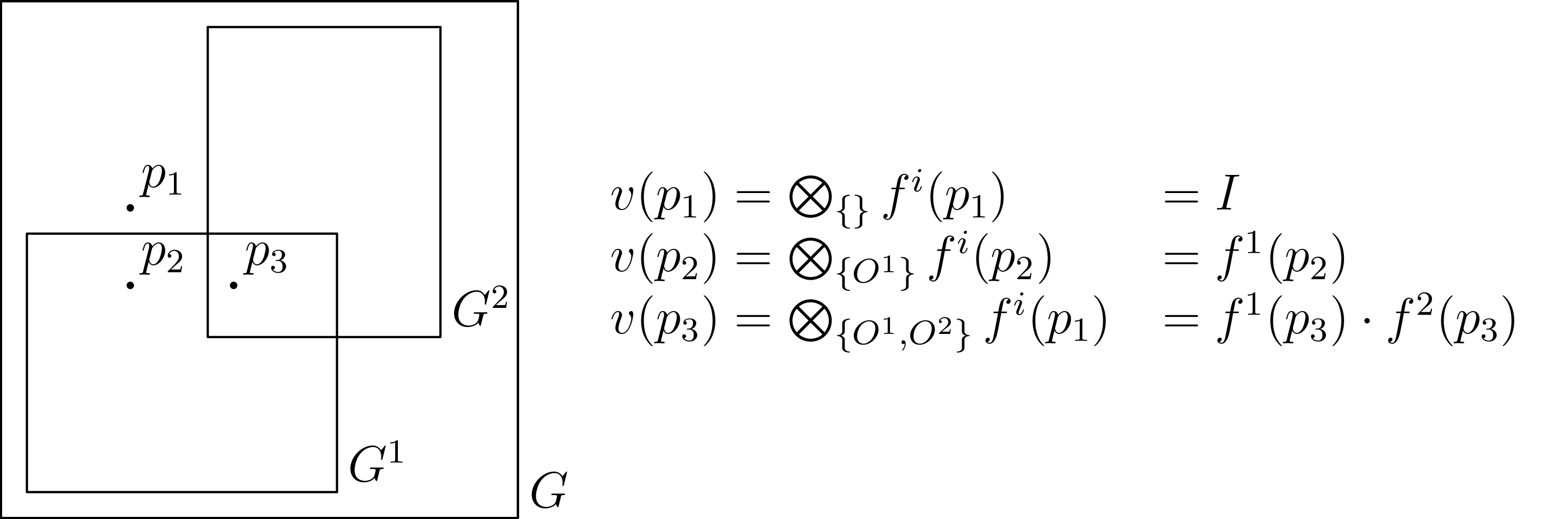}
  \caption{
    An exemplary problem setup in line with the formal problem definition:
    A set of components $O=\{O^1, O^2\}$ which is in the process of being rasterized on a grid $G$.
    On the right side, the explicit calculation of the three exemplarily chosen sample points $p_1, p_2, p_3$ as the combination of voxel values sampled from the components is shown.
}\label{fig:example}
\end{figure}

\textit{The Rasterization stage} $r\colon \Omega \rightarrow \mathcal{V}$ maps the set of components to a volume $v \in \mathcal{V}$ which is modeled as a function on the voxel grid: $v\colon G \rightarrow V$.
For a given point $(p_x, p_y, p_z)$, the voxel value $v(p_x, p_y, p_z)$ is defined as the combination of function values $f^i$ of those components $O^i$ whose bounding box contains $(p_x, p_y, p_z)$:
\begin{align}
  \label{eq:rasterization}
  v(p_x,p_y,p_z) = \underset{{\{O^i \in O \,|\, (p_x,p_y,p_z) \in G^i\}}}{\bigotimes} f^i(p_x,p_y,p_z)
\end{align}
If a voxel is not covered by any bounding box, the voxel value $v(p_x,p_y,p_z)$ is equal to identity element of $\bigotimes$.
The interplay of component bounding boxes, functions and the resulting volume is illustrated in \autoref{fig:example}.
In practice, voxel-wise noise may additionally be applied to $v(p_x,p_y,p_z)$, but this is usually inexpensive and not considered further here.
Other, non-voxel local sources of image degradation (e.g.\ blurring, implemented via convolution with a kernel $g$) may either be implemented directly in component functions as $f_i := g * f_i$ with a slightly extended bounding box or as a postprocessing pass on the generated volume.

\section{Rasterization Methods}
\label{sec:method}

This section will describe algorithms that implement the rasterization procedure, i.e.\ the computation of Equation (\ref{eq:rasterization}), defined in the previous section.
We first introduce two methods inspired from computer graphics and spatial indexing.
Then, the novel \textit{nested sweeps} algorithm, combining advantages of the previous two methods is presented and discussed, including an analysis of run time behavior.

\subsection{Component-Order Rasterization}
One way to solve the problem of rasterization of components as described above uses a method inspired by the rasterization of primitives (usually triangles) in computer graphics.
There, the screen or volume is usually first cleared, i.e.\ initialized with a known value, before triangles may be written into the subregions of the screen/volume that contain them~\cite{zhang2018scanline}.
One method to rasterize individual triangles is to compute the minimal axis aligned bounding box and for each pixel/voxel in the bounding box check whether it is within or outside of the triangle~\cite{pineda1988rasterization}.
This procedure can be adapted to the problem of model rasterization as defined above:
As a first step, the output volume is initialized with the identity value for $\bigotimes$.
Then, one component after the other, all voxels of the bounding box are iterated over, at each voxel combining the previous output volume value with the component function value.
Algorithm~\ref{alg:bb_focused} describes this procedure in pseudo-code.

\begin{algorithm}[t]
    \caption{Component-Order Rasterization}
    \hspace*{\algorithmicindent} \textbf{Input:} Components $O = \{O^1, \ldots, O^n\}$ \\
    \hspace*{\algorithmicindent} \textbf{Output:} Volume $v(p_x,p_y,p_z) \in V$
    \label{alg:bb_focused}
\begin{algorithmic}[1]
    \For{$p_z \in \{1,\ldots, d_z\}$}
        \For{$p_y \in \{1, \ldots, d_y\}$}
            \For{$p_x \in \{1, \ldots, d_x\}$}
            \State $v(p_x,p_y,p_z) = I$ \Comment{Initialization}
            \EndFor
        \EndFor
    \EndFor
    \State $O = \underset{O^i: \, (a^i_z d_y + a^i_y)d_x+ a^i_x}{\textrm{sort}}\,O$ \Comment{Optional sorting step}
    \For{$O^i \in O$}
      \For{$p_z \in \{a^i_z,\ldots, b^i_z\}$}
          \For{$p_y \in \{a^i_y, \ldots, b^i_y\}$}
              \For{$p_x \in \{a^i_x, \ldots, b^i_x\}$}
              \State \begin{varwidth}[t]{\linewidth}
                $v(p_x,p_y,p_z) =$ \par $\hskip\algorithmicindent \hskip\algorithmicindent \hskip\algorithmicindent v(p_x,p_y,p_z) \otimes f^i(p_x,p_y,p_z)$
                \end{varwidth}
              \EndFor
          \EndFor
      \EndFor
    \EndFor
\end{algorithmic}
\end{algorithm}

However, while this method works well in the 2D case where the overall size of the output data is smaller than main memory, it has drawbacks when applied to large volumetric images:
Due to caching effects on multiple levels (CPU caches, operating system's page cache for block devices and possibly further hardware caches integrated into hard drives) accessing $v$ at two locations in succession that are close in memory may result in a far lower latency for the second access than when accessing values at two locations that are far apart.
At least for hard disk drive accesses and often in general, a higher latency also implies a reduction of throughput.
Here we assume that when stored on disk the volume is linearized in order $z$-$y$-$x$, i.e.\ the memory distance of $(p_x,p_y,p_z)$ to $(p_x+1,p_y,p_z)$ is 1 (size of an element of $V$).
To $(p_x,p_y+1,p_z)$ the distance is $d_x$ while to $(p_x,p_y,p_z+1)$ it is $d_x d_y$.
In order to avoid these cache misses another variant of this method includes a sorting step according to the linear memory position of the lower left bottom corner of their bounding box.
In other words, this variant still operates in component-order, but the components are arranged in a way that is closer to the memory-order of voxels.
Overall this may also increase the run time due to the additional sorting step, but is beneficial in practice as demonstrated in the experimental evaluation (see \autoref{sec:experiments}).

\subsection{Voxel-Order Rasterization with Spatial Indexing}
\label{sec:rstar-tree}
Another rasterization approach is derived from the voxelization methods from volumetric computational solid geometry (VCSG)~\cite{liao2002csggpu,koc2004fpga} which generate the volume slice-by-slice:
For each voxel in a slice it is determined if they are in the interior or exterior of the model.
Since in this formulation each voxel interacts with all components of the model the process is usually accelerated using hardware parallelism.
This approach has the advantage of improving the locality of accesses to the output volume, but still has scalability issues if the number of components is large.
In the variant considered here (Algorithm \ref{alg:rstar-tree}) querying of components for a given voxel is therefore made efficient by using a spatial indexing structure.
Below, we assume an R$^*$ tree as the indexing structure~\cite{beckmann1990rstartree}, but other choices are possible~\cite{king1994tvtree,berchtold1996xtree,rtree2017shekhar}.
In contrast to the component-order rasterization, each voxel value is written exactly once and each final voxel value is determined immediately before writing by querying all objects whose bounding box contains the voxel position.
By iterating over all voxels in the order $z$-$y$-$x$, i.e.\ memory-order, the distance in memory between two voxels that are written after another is always 1.
This maximal spatial-locality during write operations is especially important when writing to rotating hard drives (HDDs), see \autoref{sec:experiments}.
Further, the structure of Algorithm~\autoref{alg:rstar-tree} also enables streaming processing of the resulting volume.
This is the case for this and the nested sweeps method (presented in the next section).
As a result, the created volume can be processed without writing it to disk, if a downstream method (e.g.\ a newly image processing method to be evaluated) supports this.
In general, this is desirable, since the volume generation method may actually be bottlenecked by IO (see \autoref{sec:experiments}).


\begin{algorithm}[t]
    \caption{Voxel-Order rasterization with Spatial Indexing}
    \hspace*{\algorithmicindent} \textbf{Input:} Components $O = \{O^1, \ldots, O^n\}$ \\
    \hspace*{\algorithmicindent} \textbf{Output:} Volume $v(x,y,z) \in V$
    \label{alg:rstar-tree}
\begin{algorithmic}[1]
    \State $T = \textrm{build\_index } O$
    \For{$p_z \in \{1,\ldots, d_z\}$}
        \For{$p_y \in \{1, \ldots, d_y\}$}
            \For{$p_x \in \{1, \ldots, d_x\}$}
            \State $v(p_x,p_y,p_z) = \underset{O^i \in \textrm{find}_T(p_x,p_y,p_z)}{\bigotimes} f^i(p_x,p_y,p_z)$
            \EndFor
        \EndFor
    \EndFor
\end{algorithmic}
\end{algorithm}

\subsection{Nested Sweeps Rasterization}
While the reduction of memory-/disk-access related cost of the voxel-order method using spatial indexing is useful in principle, its impact in practice is reduced due to the relatively high cost of finding components in the tree at every voxel (see \autoref{sec:experiments}).
The nested sweeps algorithm proposed in this paper and outlined in Algorithm~\ref{alg:proposed} reduces the required work at every voxel by keeping track of \textit{active} components in sweeps on each level of the three nested loops.
The outer loop implements a single sweep of all $z$-slices of the volume.
Then, for each slice a sweep over all $y$-lines within the slice is performed where each line is processed using a final sweep in $x$-direction.

\begin{algorithm}[t]
    \caption{Nested Sweeps Rasterization}
    \hspace*{\algorithmicindent} \textbf{Input:} Components $O = \{O^1, \ldots, O^n\}$ \\
    \hspace*{\algorithmicindent} \textbf{Output:} Volume $v(x,y,z) \in V$
    \label{alg:proposed}
\begin{algorithmic}[1]
    \State $Q_z = \underset{O^i: \, a^i_z}{\textrm{sort}}\,O$
    \State $C_z = \{\}$
    \For{$p_z \in \{1,\ldots, d_z\}$}
        \State $C_z = \{O^i \in C_z \,|\, b^i_z \geq p_z\} \cup \underset{O^i:\, a^i_z = p_z}{\textrm{pop}} Q_z$
        \State $Q_y = \underset{O^i: \, a^i_y}{\textrm{sort}}C_z$
        \State $C_y = \{\}$

        \For{$p_y \in \{1, \ldots, d_y\}$}
          \State $C_y = \{O^i \in C_y \,|\, b^i_y \geq p_y\} \cup \underset{O^i:\, a^i_y = p_y}{\textrm{pop}} Q_y$
          \State $Q_x = \underset{O^i: \, a^i_x}{\textrm{sort}}C_y$
          \State $C_x = \{\}$

            \For{$p_x \in \{1, \ldots, d_x\}$}
                \State $C_x = \{O^i \in C_x \,|\, b^i_x \geq p_x\} \cup \underset{O^i:\, a^i_x = p_x}{\textrm{pop}} Q_x$
                \State $v(p_x,p_y,p_z) = \underset{O^i \in C_x}{\bigotimes} f^i(p_x,p_y,p_z)$
            \EndFor
        \EndFor
    \EndFor
\end{algorithmic}
\end{algorithm}

\begin{figure*}[t]
  \centerline{\includegraphics[width=0.75\textwidth]{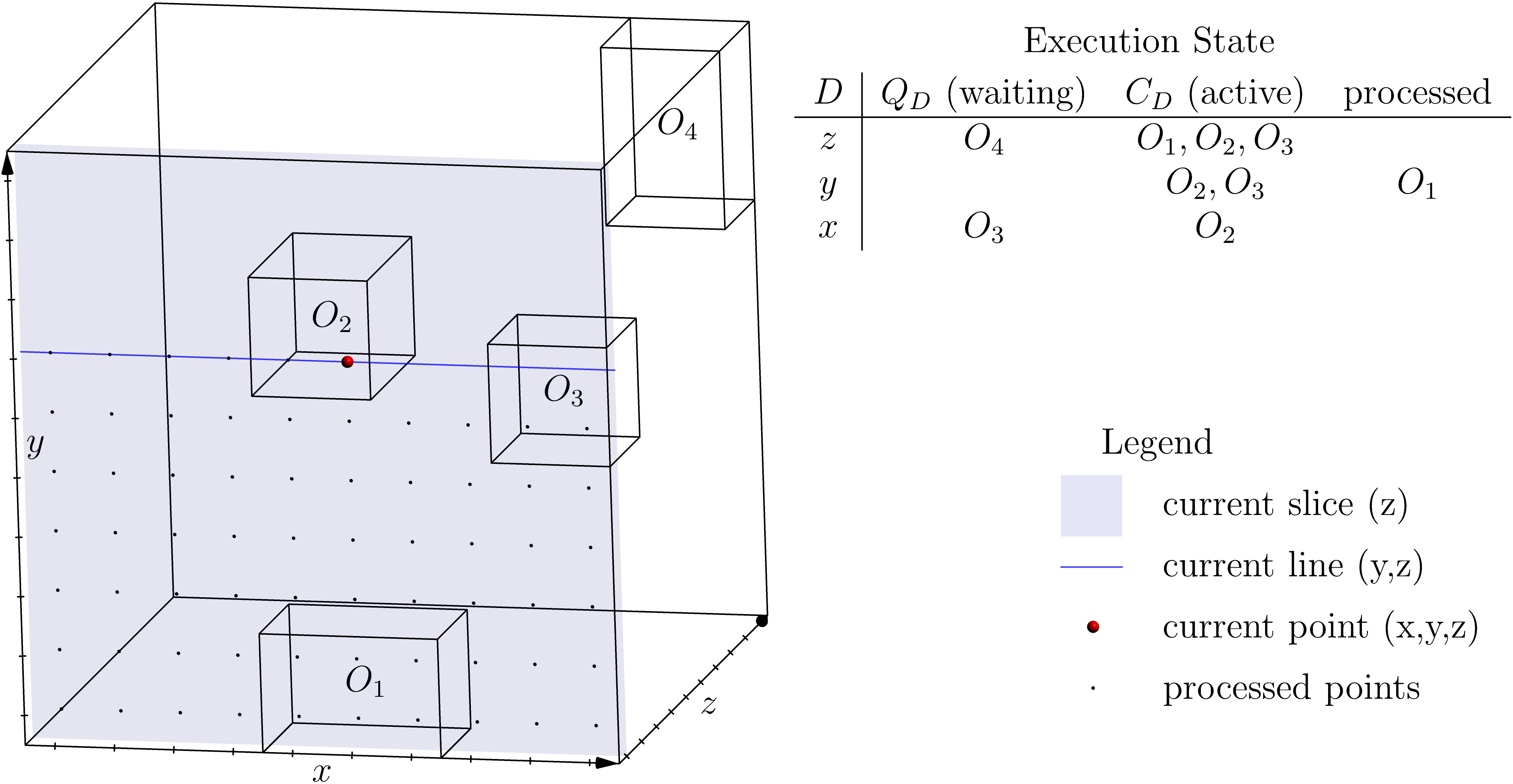}}
  \caption{
    An illustration of the processing schema of the \textit{nested sweeps} algorithm:
    The rasterization process of components $O_1$, $O_2$, $O_3$ and $O_4$ is currently in the middle of a line of the first slice.
    Since $O_1$, $O_2$ and $O_3$ all intersect with the first slice, they are currently part of $C_z$, while $O_4$ lies within the region of larger $z$ values and is thus still in $Q_z$.
    In the next inner loop, only $O_2$ and $O_3$ intersect with the current line (defined by the current values for $y$ and $z$) and are active (element of $C_y$).
    $O_1$ has already been processed completely and thus removed from $C_y$ previously.
    Within the current line, only $O_2$ contains the current point and its function will thus be used to construct the voxel value at the current position.
    $O_3$ is still waiting to become active in $Q_x$.
    It is also apparent that the only components that are considered at any stage of the computation (e.g.\ in the $y$-loop) are those, that are active in the next outer loop (e.g.\ the $z$-loop).
}\label{fig:method}
\end{figure*}

The sweep in dimension $D$ itself is implemented as follows:
As initialization, the set of components to be processed (active in the next outer sweep) is sorted according to the beginning of their bounding box $a^i_D$ and then used to initialize a queue $Q_D$.
Sweeping along dimension $D$, components are removed from the front of the queue (using pop) while $a^i_D$ matches the position $p_D$ (i.e.\ if it matches the loop index) and added to the set of active components $C_D$.
Additionally, components that have become inactive in this step of the sweep (e.g.\ if the end of the bounding box $b^i_D$ is larger than the loop index) are removed from the set.
For outer sweeps ($D \in \{z,y\}$), $C_D$ is used to initialize the queue for the next dimension.
In the innermost sweep, $C_x$ always contains components $O^i$ that are active in \textit{all} dimensions.
Hence, the current voxel is part of the bounding box $(x,y,z) \in G^i$ and all $O^i \in C_x$ are sampled to compute $v(x,y,z)$.
The algorithm is illustrated in \autoref{fig:method}.

Similar to the spatial indexing method, the computation of a single voxel value is self-contained in a single iteration of the innermost sweeping loop (line 13) and voxels are produced linearly in $z$-$y$-$x$ order.
As a consequence, the nested sweeps algorithm is suitable for streaming volume generation.

\subsection{Run Time Analysis}

In this section we will analyze and discuss the run time behavior of the methods above.
For this we assume that the cost of $v$, $f^i$ and $\otimes$ is constant.
For the component-order rasterization (Algorithm~\ref{alg:bb_focused}) we obtain a run time of $\AsO(|G| + \sum_{O^i \in O} |G^i| + n)$ where the former part of the sum accounts for the initialization and the central  and latter part for the rasterization of all components.
When including a search step, this changes to $\AsO(|G| + \sum_{O^i \in O} |G^i| + n\log n)$.

In comparison to the component-order rasterization, the increased memory/disk locality of the spatial indexing based method (Algorithm~\ref{alg:rstar-tree}) comes at the price of higher computational costs:
The initial construction of the indexing structure depends on the number of components $n$.
Specifically, for an R$^*$ tree it is log-linear in $n$.
Additionally, for each point in the volume grid $G$ the set of active components (whose bounding boxes intersect the query point) have to be determined, which is once again naturally dependent on $n$ and possible in logarithmic time in $n$ for the R$^*$ tree.
Finally, the construction and combination of component function values $f^i$ is proportional to the total number of voxels in all components, since $f^i$ is queried for each point in $|G^i|$.
Thus, in sum the run time is within $\AsO(|G|\log n + \sum_{O^i \in O} |G^i| + n \log n)$.

The run time analysis for the nested loop method (Algorithm~\ref{alg:proposed}) is somewhat more complicated.
In total, $|G|$ voxels are written and components sampled at $\sum_{O^i \in O} |G^i|$ locations (line 13).
The removal of now inactive components from $C_D$ (lines 4, 8, 12) is always strictly less costly than the subsequent or previous sorting or sampling operation.
Each component becomes active in the $x$ dimension (line 12, i.e.\ in the innermost loop) exactly $(b^i_y-a^i_y)(b^i_z-a^i_z) \leq |G^i|$ times.
The removal from other queues (lines 4, 8) is even less frequent.

Remaining are the sort operations (lines 1, 5, 9).
The initial sorting of components to construct $Q_z$ (line 1) is within $\AsO(n \log n)$.
All subsequent (individual) sorting operations (lines 5 and 9) sort subsets of $O$ and are thus already within $\AsO(n \log n)$.
However, they are executed more than once.
Let $L^{p_z}$ and $S^{p_y,p_z}$ be the number of component bounding boxes active simultaneously in slice $p_z$ and in line with coordinates $(p_z,p_y)$, respectively:
\begin{align}
  S^{p_z} &= |\{ O^i \in O \,|\, p_z \in \{a_z^i, \ldots, b_z^i\}\}|\nonumber\\
  L^{p_y,p_z} &= |\{ O^i \in O \,|\, p_z \in \{a_z^i, \ldots, b_z^i\}\ \land p_y \in \{a_y^i \ldots, b_y^i\}\}|\nonumber
\end{align}
In other words, we have $S^{p_z} = |C_z|$ in the iteration $p_z$ of the outer loop and $L^{p_y,p_z}=|C_y|$ in the iterations $p_y$ and $p_z$ of the two outer loops.
It is immediately obvious that $\forall p_y,p_z: L^{p_y,p_z} \leq S^{p_z} \leq n$.
The total cost of sorting of active slices (line 5) is within $\AsO(\sum_{p_z=1}^{d_z} S^{p_z} \log S^{p_z})$ while the cost of sorting active lines (line 9) is $\AsO(\sum_{p_z=1}^{d_z} \sum_{p_y=1}^{d_y} L^{p_y,p_z} \log L^{p_y,p_z})$.

In total, this results in an asymptotic run time of $\AsO(|G| + \sum_{O^i \in O} |G^i| + n\log n + \sum_{p_z=1}^{d_z} S^{p_z} \log S^{p_z} + \sum_{p_z=1}^{d_z} \sum_{p_y=1}^{d_y} L^{p_y,p_z} \log L^{p_y,p_z})$.
Obviously, the exact cost is highly dependent on the exact configuration of components to be rasterized.
As an example, in a bad (and likely unrealistic) case of $n$ components that span the complete volume in $z$ and $y$ direction and only occupy one voxel in $x$ direction (i.e.\ a number of $yz$-aligned slices), the run time can be more precisely specified as $\AsO(|G| + d_y d_z n\log n)$.
A more realistic case would be cells that are loosely packed and whose bounding boxes do not overlap which implies $\sum_{O^i \in O} |G^i| \leq |G|$.
Then, assuming that an increase in $|G|$ corresponds to a larger field of view (showing more cells), we obtain a run time of $\AsO(|G| \log |G|) = \AsO(n \log n)$ since $n \in \theta(|G|)$ in this scenario.
If, on the other hand, an increase in $|G|$ models an increase in camera resolution (and thus implying a constant number of cells $n$ and bounded values for $L^{p_y,p_z}$ and $S^{p_z}$) we obtain $\AsO(|G|)$ for the run time.

In general, the above considerations show that there is a trade-off between $\sum |G^i|$ and $n$ on the model generation side:
On the one hand, it is desirable to generate components so that the total component volume $\sum |G^i|$ and thus the frequency of number of times the method is required to sample from using component function $f^i$ is minimized.
This can be achieved in some cases by splitting a component.
As an example, for vessel segments consisting of cylinders with semispheres as caps, instead of generating one component per segment (as is illustrated in \autoref{fig:motivation}), it is also possible to generate separate components for the cylinder and semispheres.
In fact, it is even possible to subdivide the cylinder along its axis into an arbitrarily large number of smaller cylinders.
Especially if the central axis of the cylinder is not aligned with any of the coordinate axes, this will result in a total reduction of the volume of the \textit{axis-aligned} bounding boxes.
On the other hand, however, a larger number of components also impacts the run time of all of the above methods negatively.
Additionally, when chosen suboptimally, a subdivision may not always result in a reduction of the total component volume due to overlap between boxes.

\section{Experimental Evaluation}
\label{sec:experiments}

In this section, we will present the results of an experimental run time evaluation of the rasterization methods described above.
The objective is to generate a cubic volume of side length $d=d_x=d_y=d_z$, with a voxel volume of $|G| = d^3$.
For this, we first generate $n$ components $O=\{O^1, \ldots, O^n\}$ with random bounding boxes.
In order to show differences between the methods as effectively as possible, the function within a bounding box is simply a random constant.
The component function values are aggregated by summation.
Let the volume of component $O_i$ be $V_i = (b^i_x-a^i_x)(b^i_y-a^i_y)(b^i_z-a^i_z)$ and let $V= \underset{i \in \{1,\ldots,n\}}{\sum} V_i$ be the total volume of bounding boxes.
The bounding box dimensions are determined by drawing a uniformly distributed integer from the range $\{1, \ldots, 2m-1\}$.
The bounding box positions are also drawn from uniform distributions, ensuring that the bounding box is completely contained within the volume.
We choose $m = (\frac{|G|e}{n})^\frac{1}{3}$ for a constant $e$.
Since all samples drawn from the distribution are independent, the expected volume is $\mathbb{E}(V_i) = m^3 = \frac{|G|e}{n}$ and $\mathbb{E}(V) = \underset{i \in \{1,\ldots,n\}}{\sum} \mathbb{E}(V_i) =  m^3n = |G|e$.
We call $e$ the \textit{expected relative component volume}.
This parameter is orthogonal to the total number of voxels in the generated volume $|G|$ as well as the number of functions $n$.
Intuitively it describes the ratio between the total number of voxels in the volumetric image and the (expected) number of voxels in bounding boxes $\sum_i |G_i|$ (thus including overlap).
Below, we vary the volume size ($|G|$), number of components ($n$) and effective relative component volume ($e$) independently and observe effects on the run time between the methods.
The volume size is observed in a range from $1000^3$ to $5000^3$ voxels which corresponds to files of 4GB to 500GB (with four bytes per voxel).
The number of components includes powers of 10 from 10 to 10000 while the expected relative component volume is either 0.01 or 0.1.
For reference, the number of vessel segments (and thus components) is 100 in a provided example configuration file for VascuSynth and 10000 in the volume generated in \autoref{sec:case_study}.
Similarly, the example configuration file results in \textit{observed} relative component volumes between 0.01 and 0.1, depending on the rasterization (i.e.\ volume size) and exact number of vessel segments.
In this experiment, we chose an R$^*$ tree with an existing, popular implementation\footnote{\url{https://github.com/georust/rstar}} for the voxel-order rasterization with spatial indexing.
In addition to the methods described above -- component-order rasterization (``$O^i$-Order``), voxel-order rasterization with spatial indexing (``Spatial Indexing``) and the nested sweeps method (``Nested Sweeps``) -- we also run benchmarks for a hypothetical best-case method (``Baseline``) which does not actually solve the problem of test data generation, but writes a volume with constant voxel value and of the specified size $|G|$ to disk.
This serves as a best-case lower bound for the run time.

All experiments were conducted on a consumer grade computer with a AMD Ryzen 7 2700X (3.7 GHz) CPU, 32GB RAM.
The volumes were either written to a Samsung NVMe SSD 960 EVO (\textit{SSD}) or a Western Digital Red 5400RPM SATA drive (\textit{HDD}).
Experiments were repeated 5 times with different random seeds and median as well as minimum and maximum times are reported below.

\subsection{Variation of Volume Size}

\begin{figure}[t]
  \begin{center}
  \includegraphics[width=0.49\columnwidth]{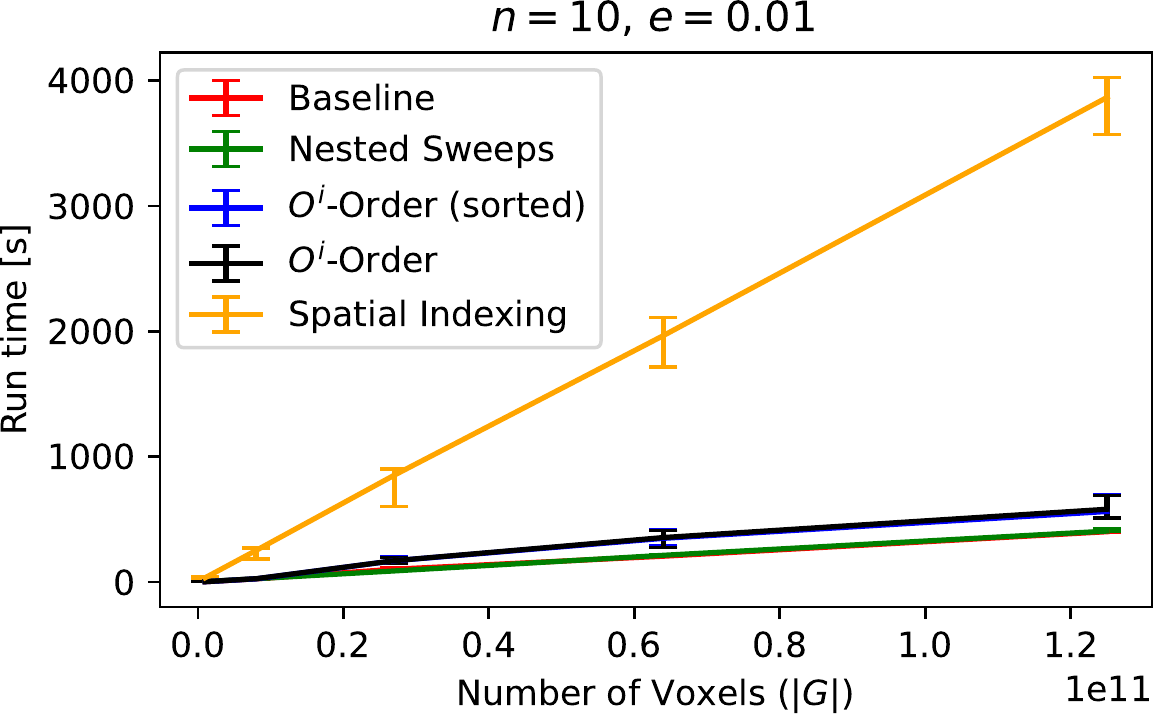}
  \includegraphics[width=0.49\columnwidth]{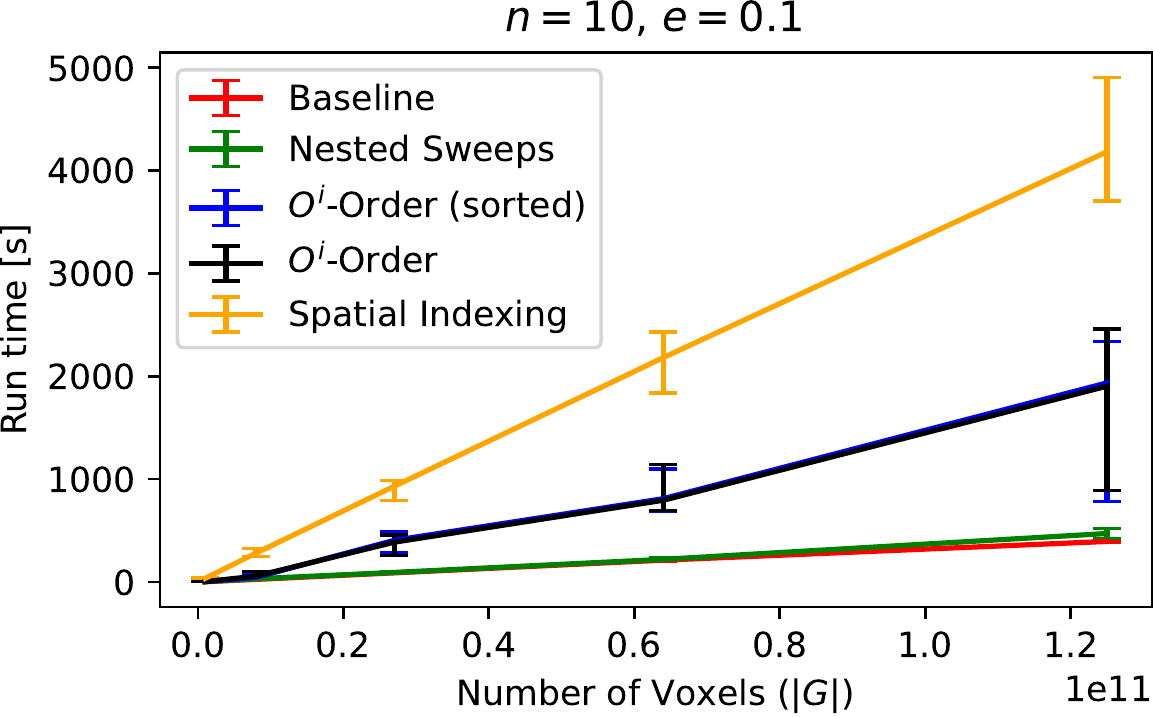}
  \\\vspace{2mm}
  \includegraphics[width=0.49\columnwidth]{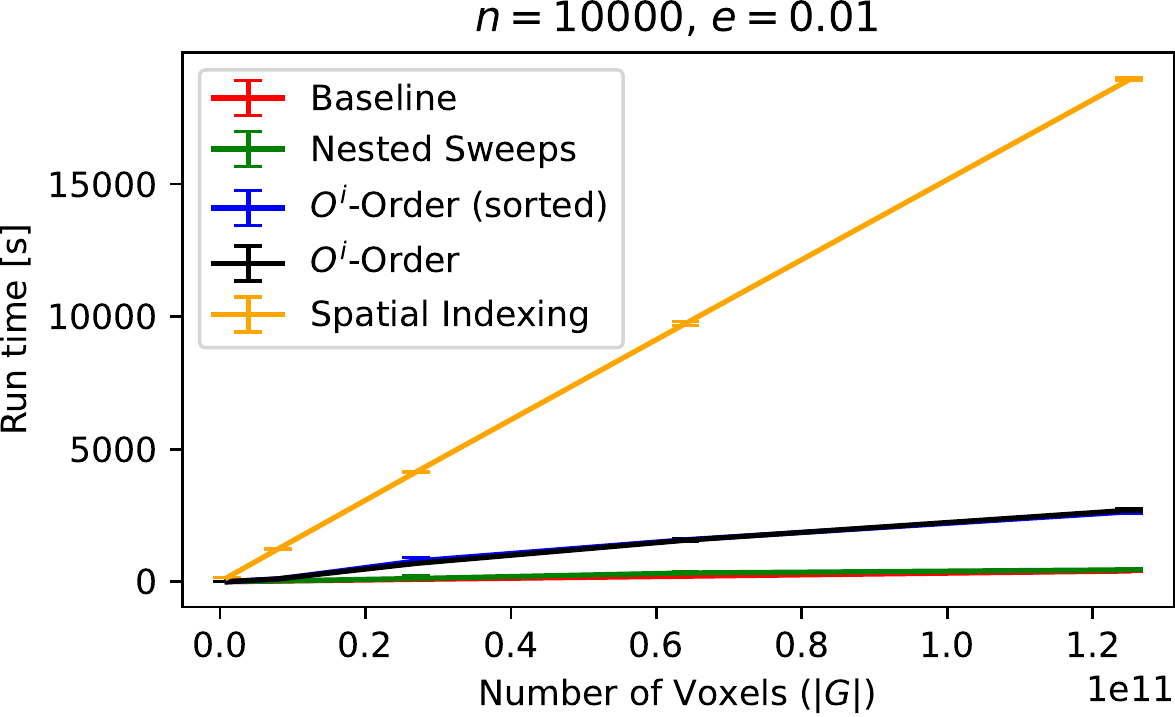}
  \includegraphics[width=0.49\columnwidth]{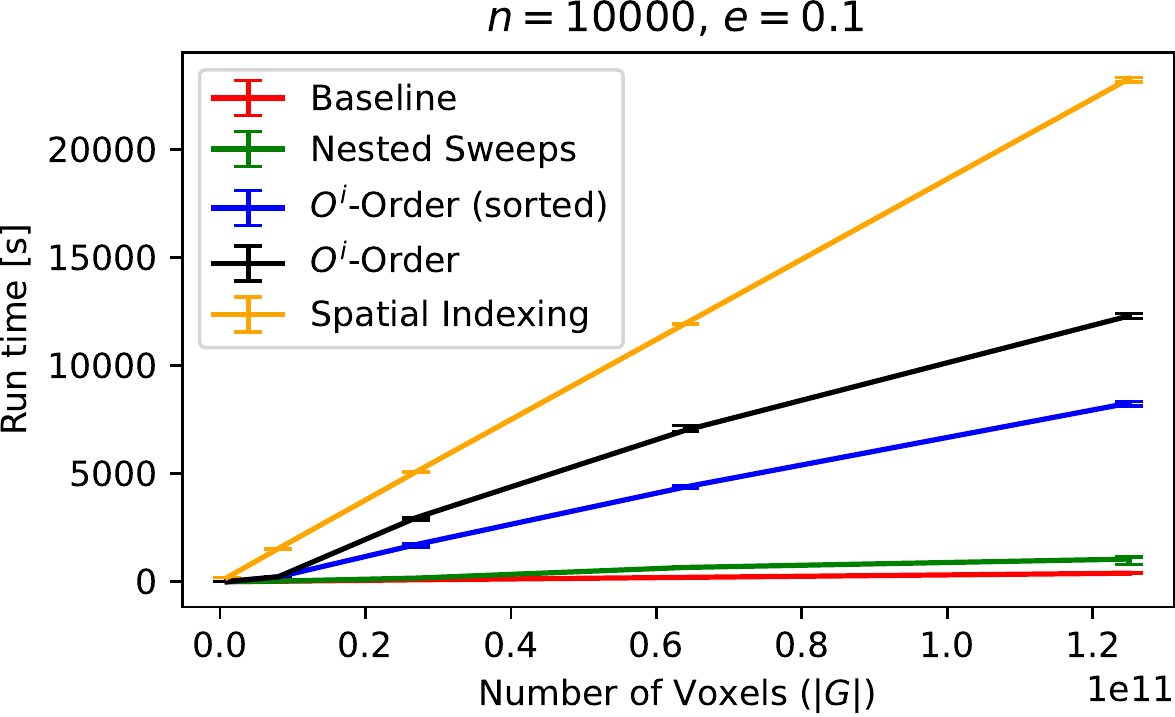}
\end{center}
\caption{Run time analysis of the nested sweeps and other methods when writing to a \textbf{solid state drive}: Each figure shows the run time for different output volume sizes. The figures show all combinations of extreme values for $n$ and $e$: $\{10, 10000\}\times\{0.01, 0.1\}$}\label{fig:size_ssd}
\end{figure}
\autoref{fig:size_ssd} shows the run time trend for an increasing volume size $|G|$ and different, but constant values for $e$ and $n$ when writing the volume to a solid state drive.
It is immediately obvious that in this case, the spatial indexing based method is significantly slower than any of the other methods.
However, as should be expected from the analysis performed previously (see \autoref{sec:method}), all methods show a (roughly) linear increase in run time with an increasing $|G|$.
Apparently, the spatial indexing based method differs from the others by a constant factor which may be explained by the increased required computational work of traversing the R$^*$ tree for each voxel.
The component-order methods exhibit an intermediate level of run time performance in this test for all depicted combinations of $n$ and $e$.
For $n=10000$ and $e=0.1$ there is a clear difference between the variants with and without sorting of bounding boxes where the variant with sorting only requires roughly 60\% of the run time compared to the variant without sorting.
This is in contrast to all other depicted configurations where no difference between variants is noticeable.
One explanation for this would be the improved locality of disk-accesses in the variant with sorting where disk-pages that contained (parts of) the previous bounding box are already loaded into memory after its rasterization and thus do not have to be loaded again or written back to disk before the rasterization of the next component.
Notable is also the increase of slope at $2000^3$ voxels which corresponds to a volume of 32GiB and is also random access memory (RAM) size of the test machine.
This further strengthens the hypothesis of problematic volume access patterns of bounding-box based methods.

The nested sweeps method performs best out of the set of the non-baseline methods.
For $e=0.01$ the run time curve is visually indistinguishable from the baseline curve in \autoref{fig:size_ssd}.
For a larger relative component size ($e=0.1$) the method is slightly slower than the baseline, but only marginally.
Specifically, for $n=10000$ and a volume of $5000^3$ voxels, the nested sweeps method with 1050s is within a factor of roughly 2.5 from the theoretical limit (397s) in terms of the mean computation time, while component-order methods are slower by a factor of 21 (sorted, 8268s) or 31 (unsorted, 12302s) and the spatial indexing based method by a factor of roughly 59 (23280s).

\begin{figure}[t]
  \begin{center}
  \includegraphics[width=0.49\columnwidth]{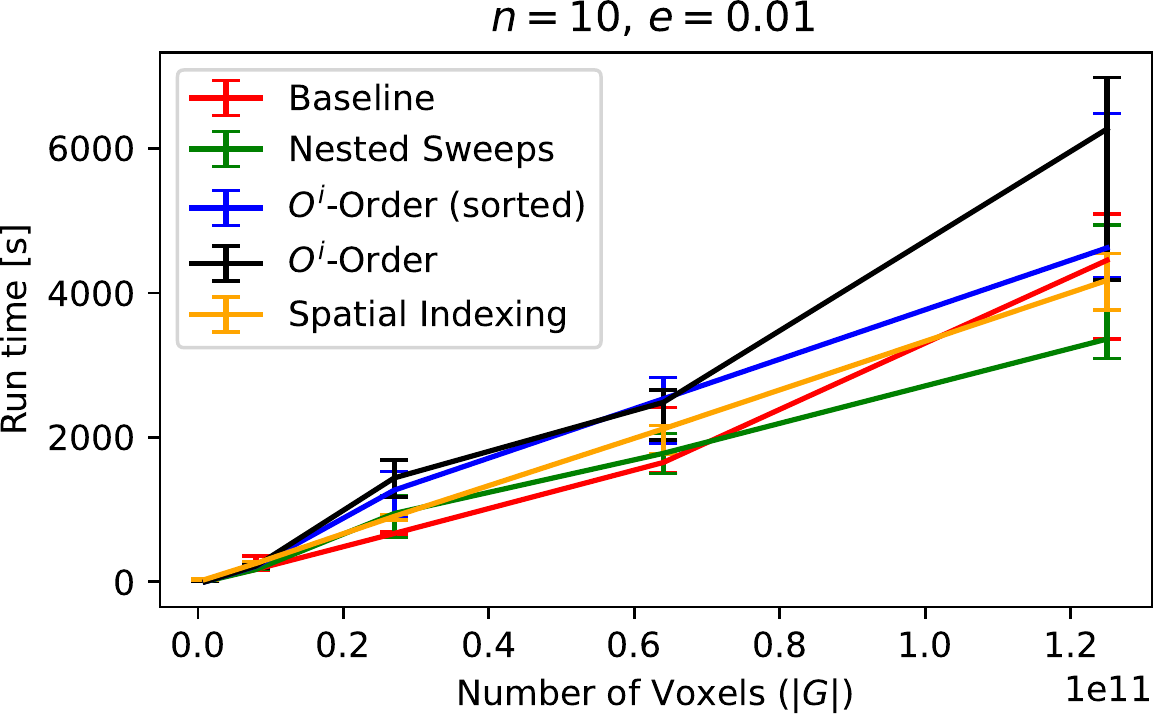}
  \includegraphics[width=0.49\columnwidth]{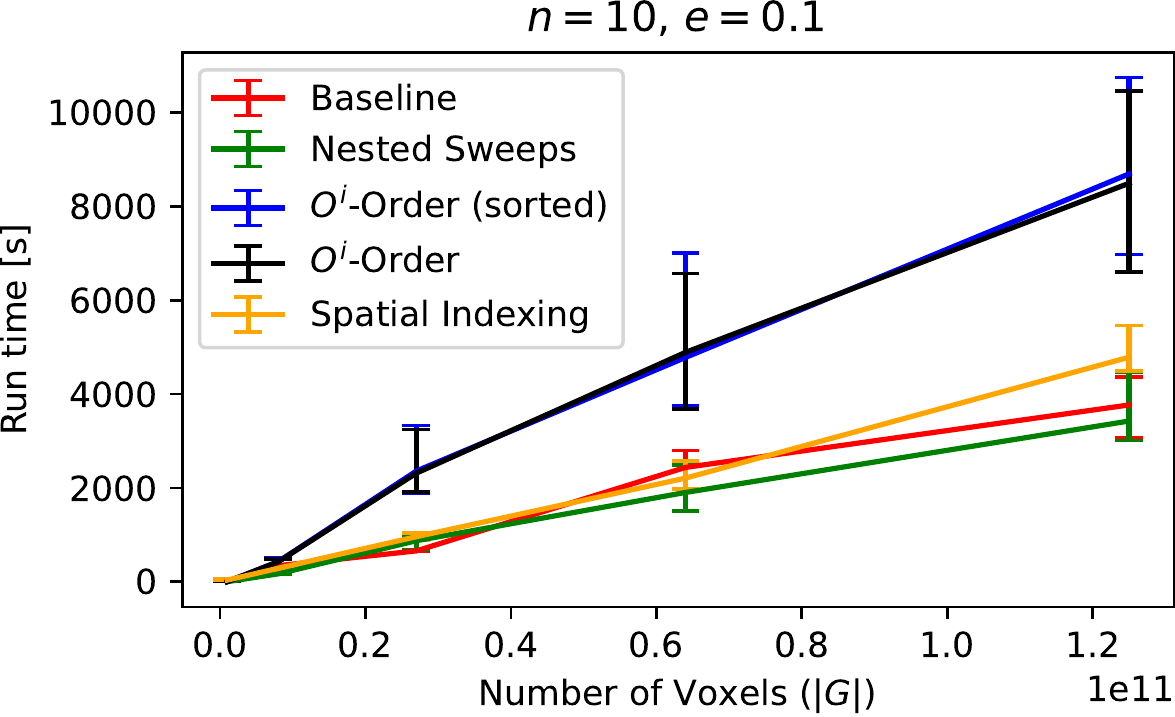}
  \\\vspace{2mm}
  \includegraphics[width=0.49\columnwidth]{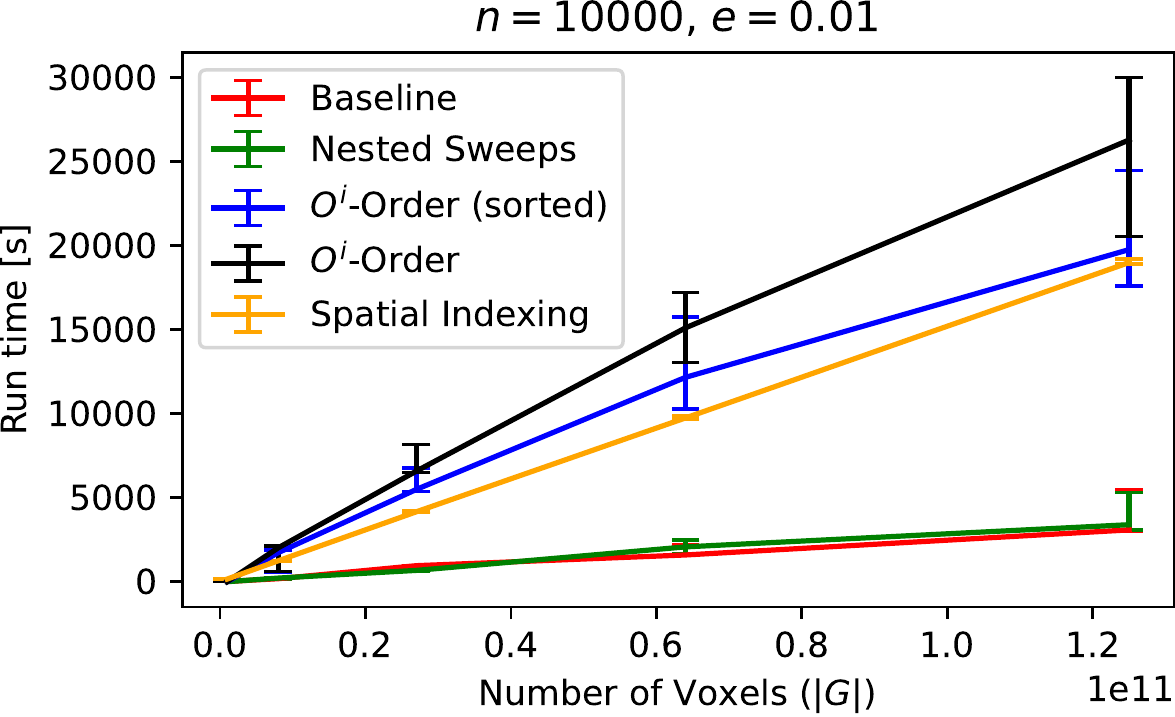}
  \includegraphics[width=0.49\columnwidth]{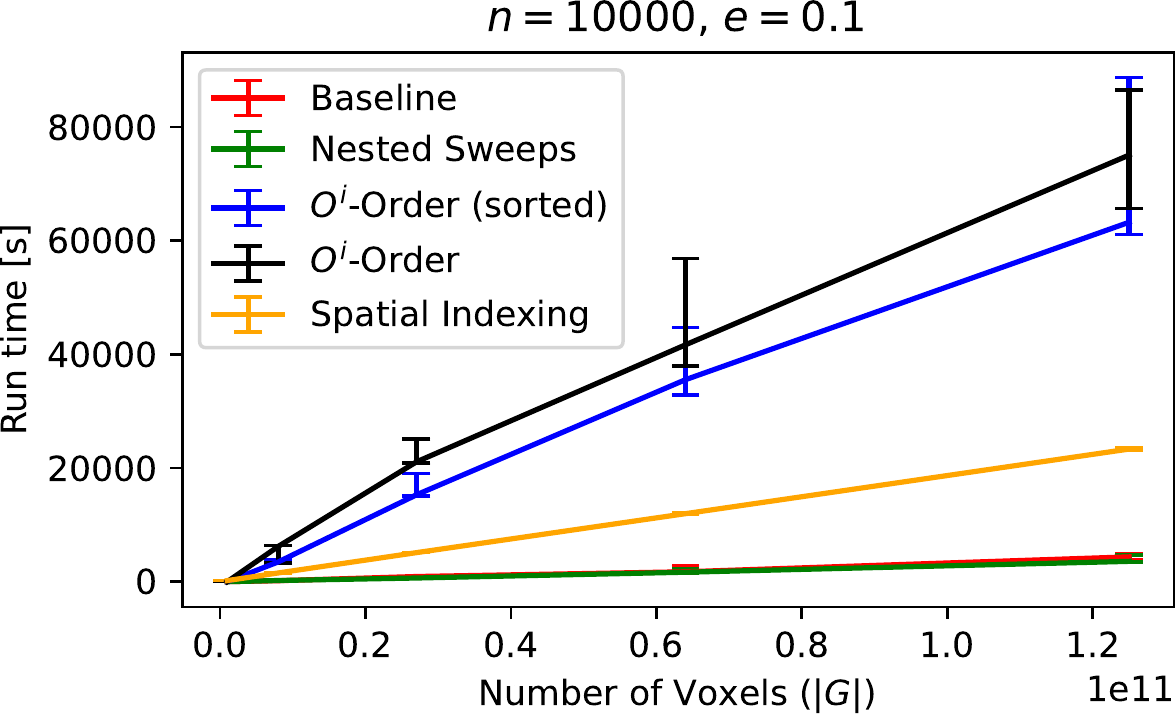}
\end{center}
\caption{Run time analysis of the nested sweeps and other methods when writing to a \textbf{hard disk drive}: Each figure shows the run time for different output volume sizes. The figures show all combinations of extreme values for $n$ and $e$: $\{10, 10000\}\times\{0.01, 0.1\}$.}
\label{fig:size_hdd}
\end{figure}

\begin{figure}[t]
  \begin{center}
  \includegraphics[width=0.49\columnwidth]{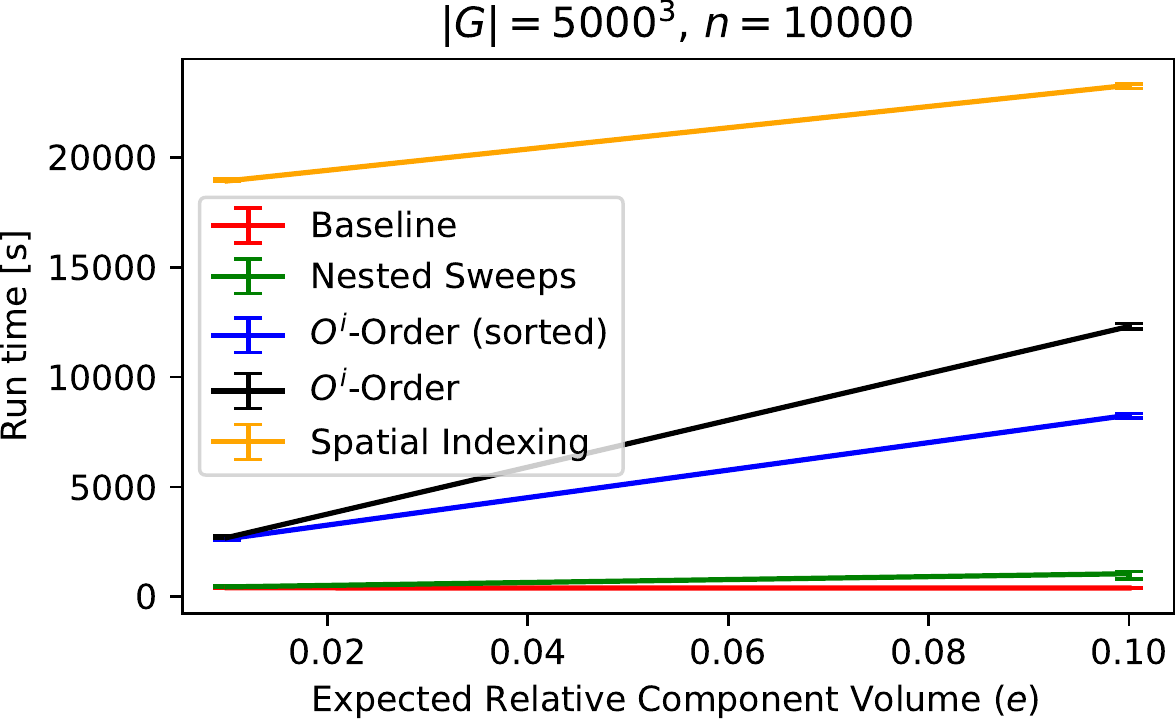}
  \includegraphics[width=0.49\columnwidth]{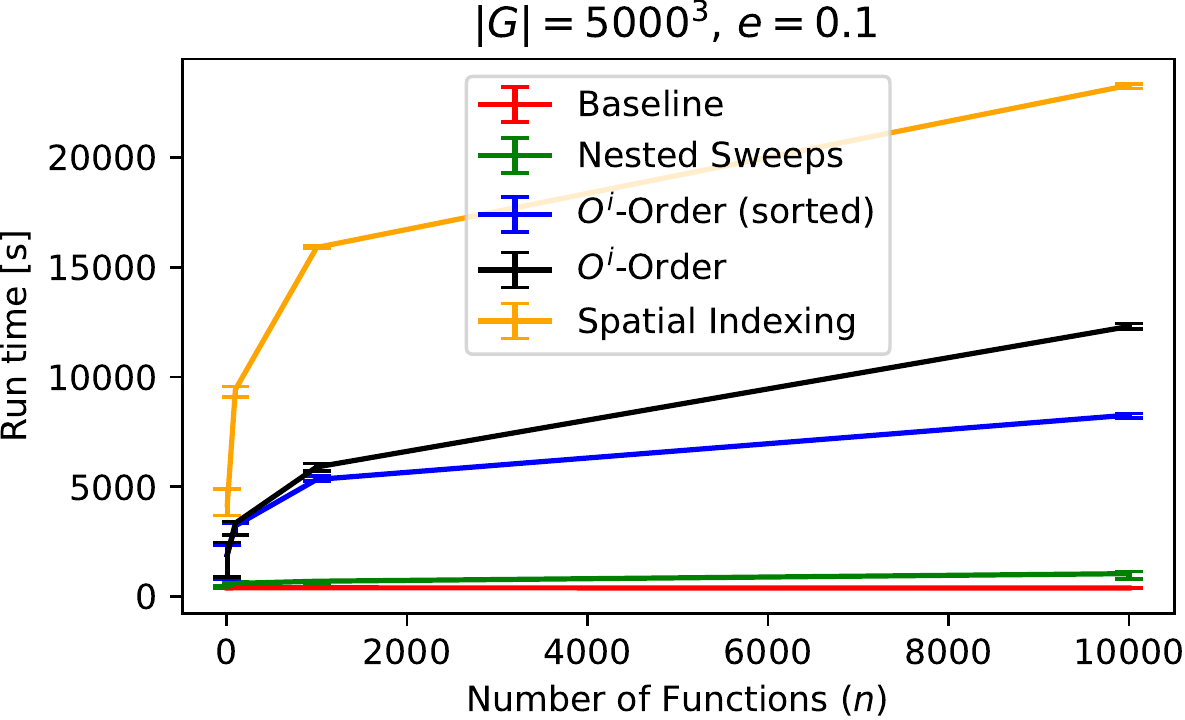}
\end{center}
\caption{Run time analysis of the nested sweeps and other methods when writing to a \textbf{solid state drive} for a fixed volume size of $|G| = 5000^3$: Left: Comparison for $e \in \{0.01, 0.1\}$, right: Comparison for $n \in \{10, 100, 1000, 10000\}$.}\label{fig:e_and_n_ssd}
\end{figure}

\begin{figure}[h]
  \begin{center}
  \includegraphics[width=0.49\columnwidth]{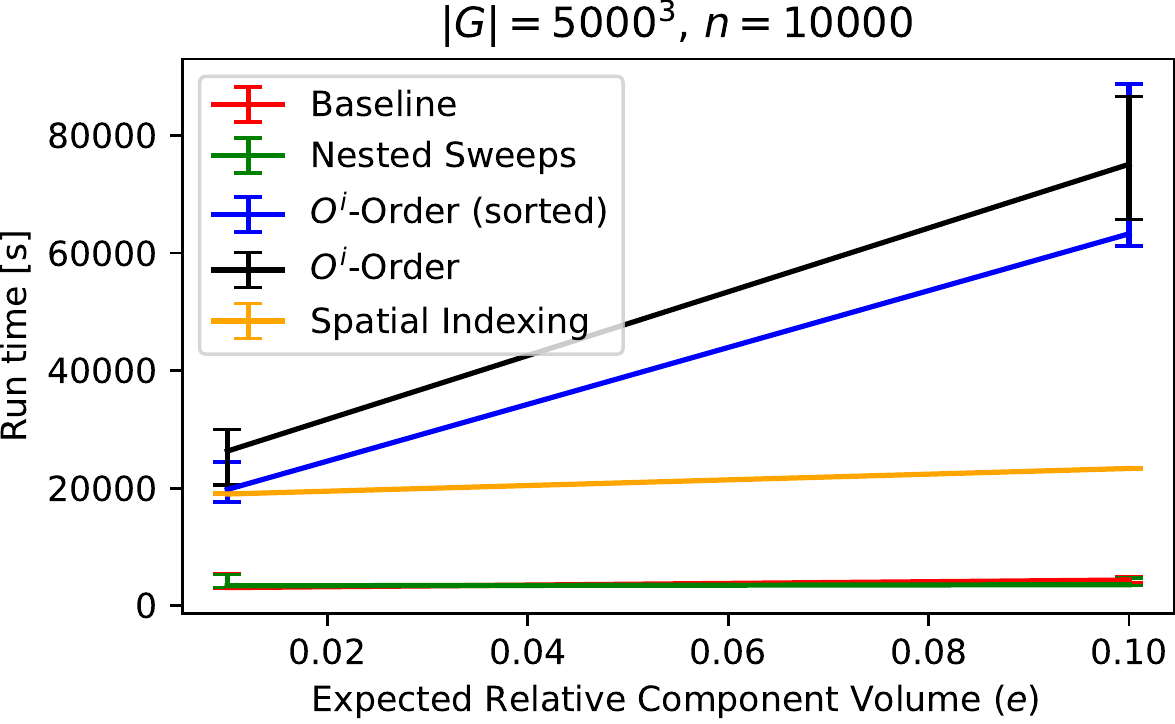}
  \includegraphics[width=0.49\columnwidth]{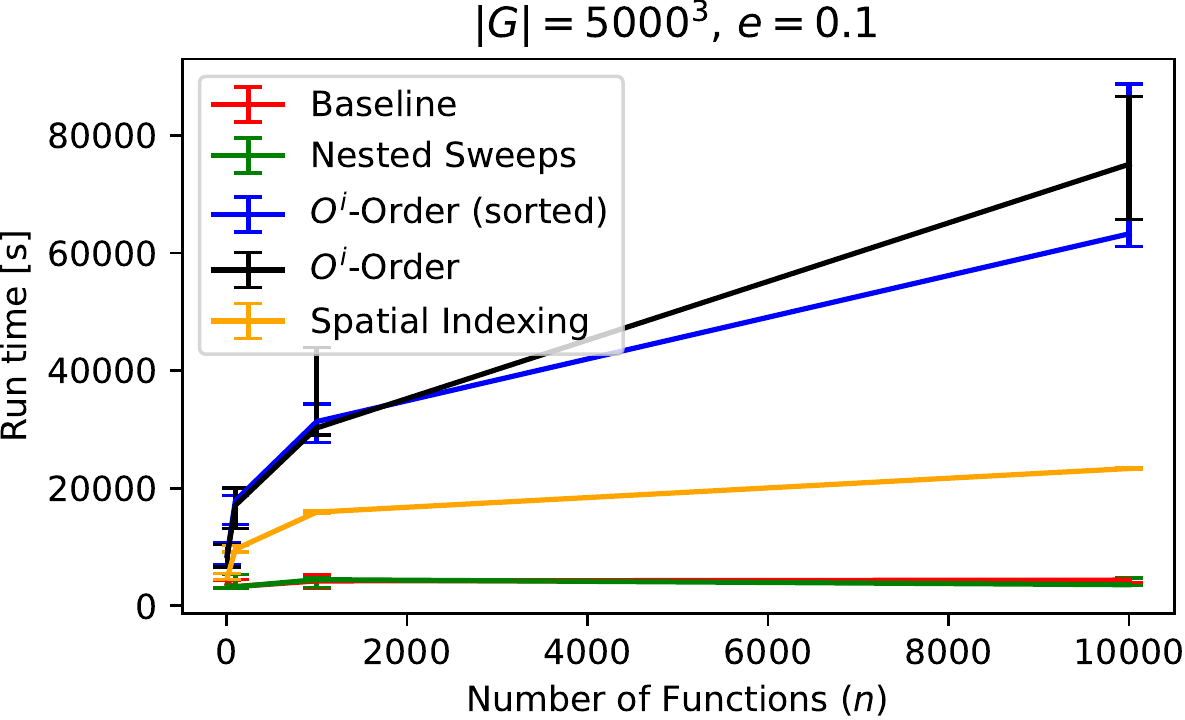}
\end{center}
\caption{Run time analysis of the nested sweeps and other methods when writing to a \textbf{hard disk drive} for a fixed volume size of $|G| = 5000^3$: Left: Comparison for $e \in \{0.01, 0.1\}$, right: Comparison for $n \in \{10, 100, 1000, 10000\}$.}\label{fig:e_and_n_hdd}
\end{figure}

\begin{figure*}
  \begin{center}
  \includegraphics[width=0.49\textwidth]{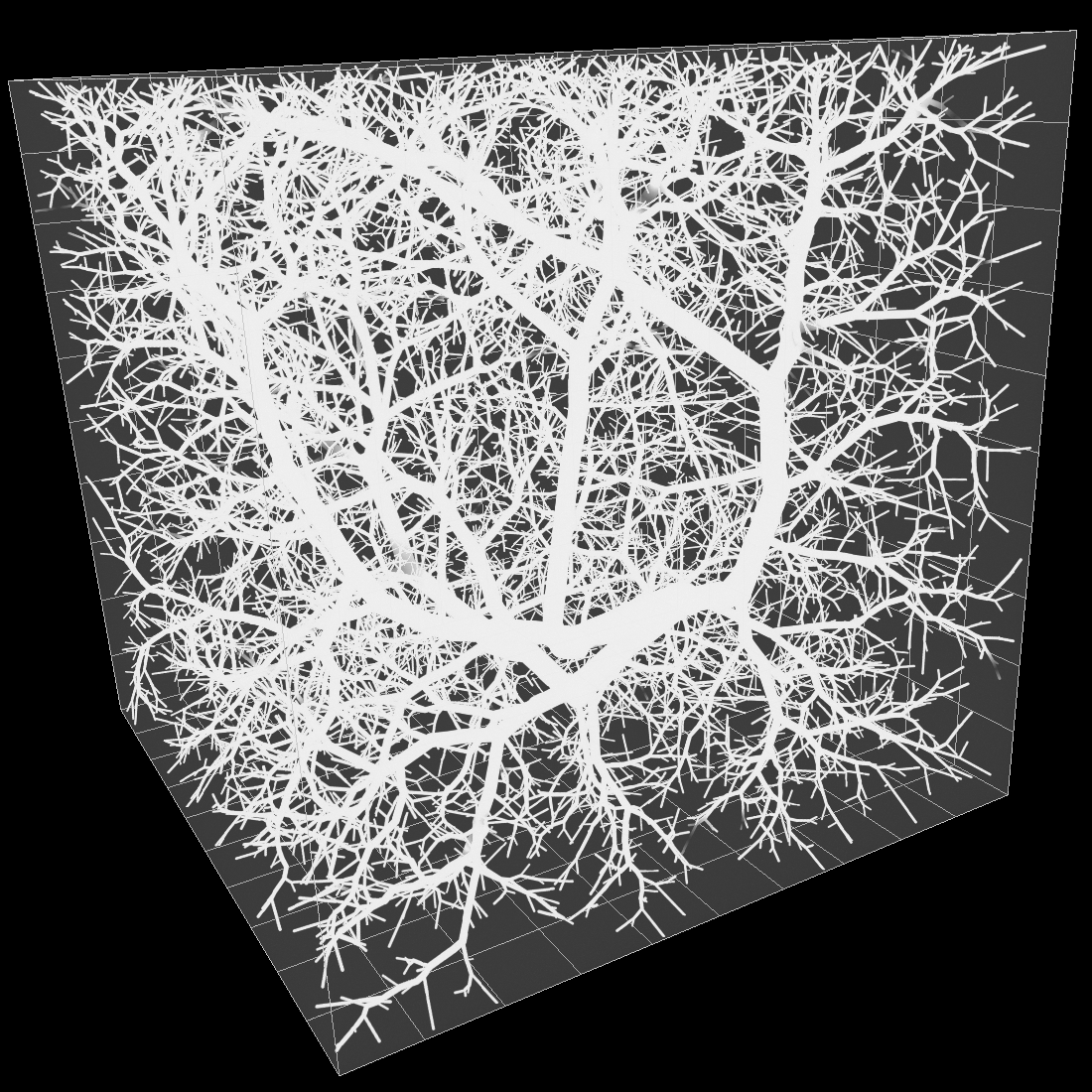}
  \includegraphics[width=0.49\textwidth]{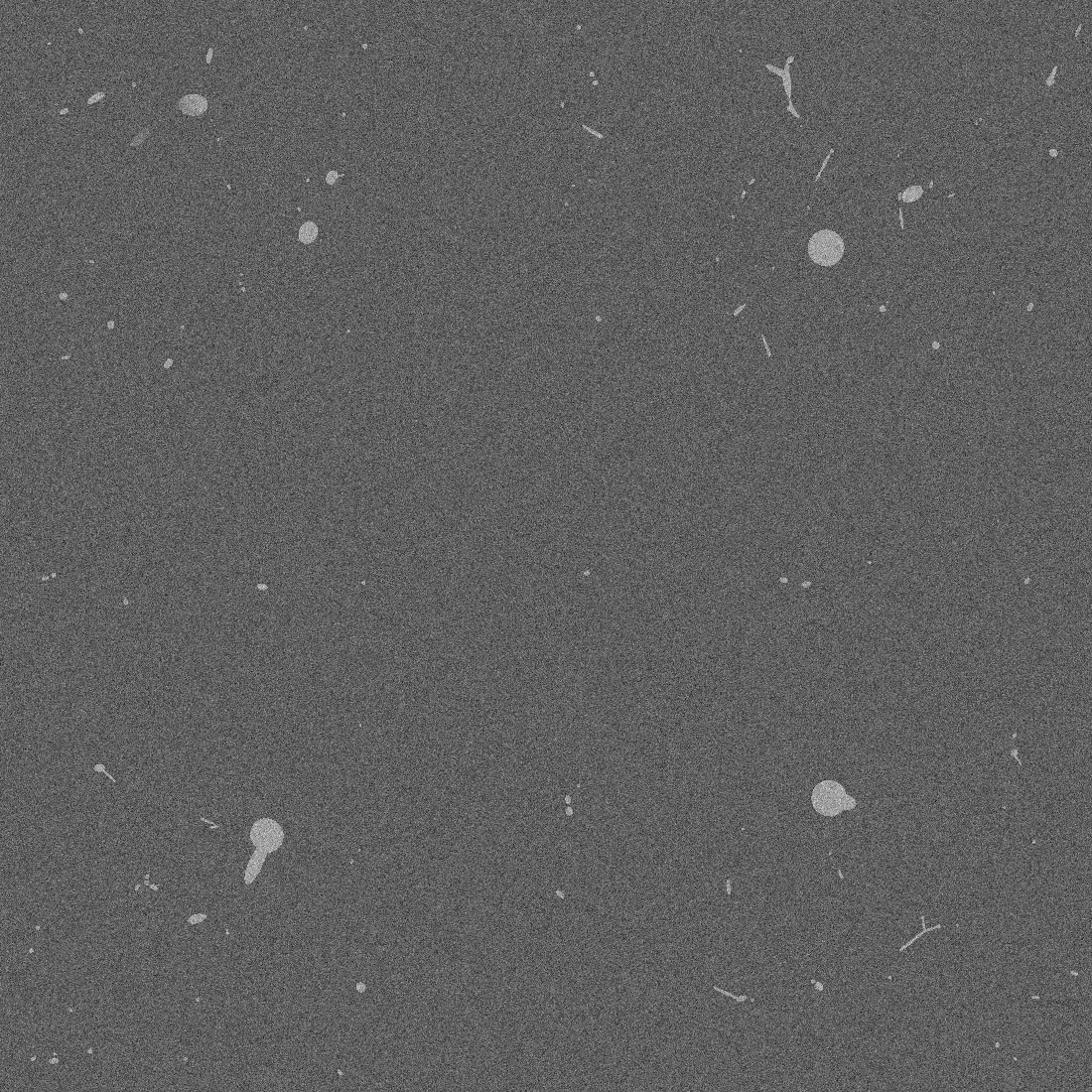}
\end{center}
\caption{
  A maximum intensity projection (left) and a single slice (right) of a $10000^3$ voxels (1TB, one byte per voxel) volume generated by the modified version of VascuSynth.
  The high resolution of the dataset allows rasterization of vessels at varying scales and thus overall a very complex vessel network structure.
  Both renderings were created using the volume rendering and processing engine Voreen~\cite{meyer2009voreen}.
}\label{fig:vascusynth_large}
\end{figure*}

\autoref{fig:size_hdd} also shows the run time trend for increasing volume sizes, but in this case when writing to a hard disk drive (HDD).
Here -- in contrast to the previous experiment when writing to an SSD -- the spatial indexing based method requires roughly the same time for completion of the task ($e=0.01$) or even less ($e=0.1$) than the component-order rasterization method.
Especially for a larger expected relative component size or a large number of components, the component-order methods (independent of sorting) are significantly slower.
This is the continuation of the trend already discussed above (in the SSD case), once again likely caused by the non-locality of disk-accesses.
In contrast to this, the nested sweeps method exhibits run time behavior that is very close to the theoretical optimal-case baseline regardless of volume size, expected relative component size or number of components:
For $|G| = 5000$, $e=0.1$ and $n=10000$, the window of observed values between baseline ([3871s, 4854s]) and nested sweeps ([3512s, 4726s]) largely overlap.
The median time for spatial indexing based method is roughly 5 times that of the baseline method while the component-order methods are slower by a factor of roughly 14 and 17, respectively.
It can thus be concluded that in this experiment, the write-throughput of the HDD is actually the bottleneck for the nested sweeps algorithm, but not for other methods.
Finally, it should be noted that -- especially for $e=0.01$ and $n=10$, i.e.\ for computationally less demanding cases -- the baseline time was actually measured to be higher than, for example, the time of the nested sweeps algorithm, which is simply measurement noise, for example due to the file system placing blocks in higher or lower throughput regions (outer or inner cylinders) of the hard drive.
This conclusion is based on the fact that the baseline method is simply the initial step of the nested sweeps and spatial indexing methods and further strengthens the hypothesis of the hard drive being the bottleneck in this setting.

\subsection{Variation of Expected Relative Function Volume and Number of Components}

\autoref{fig:e_and_n_ssd} and \autoref{fig:e_and_n_hdd} show the run time behavior of the examined methods while varying the expected relative component volume (left) and number of functions (right) on SSDs and HDDs, respectively.
It is immediately obvious that also for changes of the other two remaining experiment parameters the nested sweeps algorithm is considerably faster than other methods and that it is in fact very close to the baseline speed.
It is not surprising that the run time for bounding-box based methods increases for an increasing $e$.
Notably, this is also the case for the spatial indexing based method on the SSD, but less so on the HDD.
When increasing the number of components, unsurprisingly, all methods exhibit an increase in required run time, although the increase does not appear to be linear.
For the spatial indexing based method this is in line with the predicted (part of the) asymptotic run time that is logarithmic in $n$.

Finally it should be noted that this experiment shows that while the component-order based method is significantly faster than the spatial indexing based method when writing to an SSD, this trend is reversed when writing to a HDD.
The nested sweeps algorithm outperforms both by combining the locality of data/memory access of the voxel-order spatial indexing based method with the low computational overhead of the component-order method, and as a consequence outperforms both by a large margin independent of the target medium.
Further, when operating in a streaming fashion (i.e.\ directly passing them to the application to be benchmarked) and thus without the potential bottleneck of disk throughput, the low computational overhead of the nested sweeps method can be expected to have an even larger impact when compared to spatial indexing methods, and the component order method is not even applicable in that scenario.

\section{Case Study: Extension of VascuSynth to Large Image Data}
\label{sec:case_study}
In order to demonstrate the usefulness and applicability of the proposed framework, we modify and extend the VascuSynth test data generation software~\cite{jassi2011vascusynth} to support efficient generation of larger-than-main memory volumes.
The VascuSynth pipeline can be seen as two stages: (1) Vessel tree generation and (2) vessel image generation (cf. \autoref{fig:motivation}).
Stage (1) creates a list of vessel segments, each defined by a start and end point as well as radius.
In Stage (2) each vessel segment is modeled as the union of a cylinder with start and end points as the center of the two caps and two spheres around the start and end points.
The segment radius defines the radius of both the spheres and the cylinder.
Using these primitives, the vessel tree is rasterized before optionally applying voxel-wise noise and shadows at the mid-point of randomly selected vessel segments.
The latter is defined by a reduction of the image signal proportional to $r$ minus the distance to the mid-point (clamped to 0), where $r$ is the radius of the spherical shadow and chosen as the length of the vessel segment.

In this case study, we leave stage (1) unchanged and replace the original image generation implementation (2) which creates the whole volume in RAM first and then uses a procedure that is equivalent to Algorithm~\autoref{alg:rstar-tree} \textit{without} spatial indexing (such that find$_T$ is linear in $n$).
While this is less problematic for small datasets, this inefficient approach prevents generation of larger and especially out-of-core volumes.

For the rasterization of the vessel tree, we interpret each vessel segment (consisting of the union of two (semi-)spheres and a cylinder) as a component for Algorithm~\ref{alg:proposed}.
The bounding box is computed using the minimum/maximum of the start/end points of the segment plus/minus the radius.
The voxel value function is simply the inside-test that is defined by the geometric primitives (sphere and cylinder).
This is the same test that is already used in the original implementation.
Thus, compared to the original VascuSynth implementation, the proposed method samples the inside-test function only when a voxel is within the axis aligned bounding box of the segment geometry instead of unconditionally.

The application of shadows as described above can also be accelerated by the proposed method:
Since the signal reduction is zero outside of the radius of the shadow anyway, it does not have to be computed for all combinations of shadow and voxel in the image (as is the case in the original implementation), but instead only when a voxel is within the cubic bounding box (with side length $2r$ around the center point).
Thus the reduction factor and bounding box constitute a component as defined above and the proposed framework can be applied.
Voxel-wise noise can be applied before writing each voxel.

Using this modified version of VascuSynth we have generated a volume of $|G|=10000^3$ voxels (1TB) with voxel-wise Gaussian noise and 100 shadows.
The tree itself consists of $n=10000$ segments.
The resulting relative component volume is $e=0.092$.
\autoref{fig:vascusynth_large} shows a maximum projection rendering (left) and a single slice of the volume (right).
As can be seen, the generation of volumes of this size allows rasterization of vessel networks with vessels at multiple size scales and thus enables VascuSynth to, for example, aid research into and evaluation of multiscale image processing techniques.
The generation of a noisy image and corresponding ground truth in total required roughly one day of computation time of which 8.5 hours were required for the model generation process (which was not modified).
The rasterization of the ground truth required roughly 3.5 hours, which corresponds to a throughput of 81MB/s and thus roughly the maximum write-throughput of the HDD.
Creation of the noisy image was done in 12 hours, most of which can be attributed to the relatively slow Gaussian noise generation, as shown by profiling.
Finally it should be noted that the rasterization process -- while being far more efficient -- is functionally equivalent to the process in the original VascuSynth software, i.e.\ it produces identical outputs.
It can therefore be used as an efficient drop in replacement for the popular original implementation \cite{mou2021cs2,zeng2018automatic,titinunt2019vesselnet,wang2020tensorcut,koonjoo2021boosting} with the additional support for out-of-core volume generation.

\section{Conclusion}
In view of rapid advances in imaging technologies and the resulting need for development of out-of-core image analysis and processing methods, reliable, flexible and efficient methods for artificial dataset generation are required.
To fulfill this need, we have developed a general framework that allows common volumetric test data generation pipelines to be extended in order to allow for efficient (and optionally streaming) creation of larger-than-main memory datasets using one of the introduced volume rasterization methods.
In the experimental evaluation we have demonstrated that the novel nested sweeps algorithm outperforms two alternative approaches that are based on common practice in computer graphics and on spatial indexing structures.
The nested sweeps method combines the locality of data access and low computational overhead of the alternative methods.
As a case study, the framework has been used to successfully augment the popular vascular data generation tool VascuSynth~\cite{hamarneh2010vascusynth} with the capability to generate out-of-core volumes.

In the future we plan to apply the presented framework to other popular data generation methods and tools (e.g.\ \cite{svoboda2017mitogen, cuntz2010neuronal}) in order to increase the test data generation tool box of the large volume processing research community.
We will of course also use these tools in our own continued effort~\cite{drees2021scalable, drees2021hierarchical} to develop more out-of-core image processing methods.

The program code of the experimental evaluation\footnote{\url{https://zivgitlab.uni-muenster.de/ag-pria/large-test-data-generation-experiments}} and the modified version of VascuSynth\footnote{\url{https://zivgitlab.uni-muenster.de/ag-pria/vascusynth}} are available online.
The framework has also been used to implement large test data generation methods in the volume rendering and processing framework Voreen\footnote{\url{https://voreen.uni-muenster.de}} which was also used in \cite{drees2021hierarchical}.

\section{Acknowledgements}
This work was funded by the Deutsche Forschungsgemeinschaft (DFG) – CRC 1450 – 431460824.

\bibliographystyle{IEEEtran}
\bibliography{src}

\end{document}